\documentclass{article}

\PassOptionsToPackage{numbers,compress}{natbib}
\usepackage[preprint]{neurips_2026}

\usepackage[utf8]{inputenc} 
\usepackage[T1]{fontenc}    
\usepackage{hyperref}       
\usepackage{url}            
\usepackage{booktabs}       
\usepackage{amsfonts}       
\usepackage{nicefrac}       
\usepackage{microtype}      
\usepackage{xcolor}         

\usepackage{booktabs}
\usepackage{multirow}
\usepackage{pifont}

\usepackage{amsmath}
\usepackage{graphicx}
\usepackage{wrapfig}
\usepackage{amssymb}
\usepackage{caption}
\usepackage{booktabs}
\usepackage{amssymb}
\usepackage{caption}
\usepackage{subcaption}
\title{Hierarchical Dual-Subspace Decoupling for \\ Continual Learning in Vision--Language Models}

\author{%
Mengxin Qin\thanks{These authors contributed equally.} \quad Xiang Zhang\footnotemark[1] \quad Kun Wei\thanks{Corresponding author.} \quad Xu Yang \quad Cheng Deng\\
School of Electronic Engineering, Xidian University Xi'an 710071, China\\
}

\begin{document}

\maketitle

\begin{abstract}
Class Incremental Learning (CIL) aims to continuously acquire new knowledge while preserving previously learned information, thereby mitigating catastrophic forgetting. Existing methods primarily restrict parameter updates but often overlook their structural properties in high-dimensional spaces. From a subspace perspective, updates induced by different tasks tend to lie in multiple overlapping low-rank subspaces, leading to cross-task subspace interference and severe forgetting. To address this issue, we propose a \textbf{H}ierarchical \textbf{D}ual-\textbf{S}ubspace \textbf{D}ecoupling (HDSD) framework for Continual Learning in Vision--Language Models. Specifically, we introduce a lightweight Feature Modulation Module (FMM) that explicitly decomposes the parameter space into general and task-specific subspaces. Building upon this, we design two complementary components. First, a General Fusion Module (GFM) evaluates relative parameter changes across tasks and utilizes an adaptive threshold to capture stable, transferable knowledge. Second, a Hierarchical Learning Module (HLM) performs structured parameter decomposition via Singular Value Decomposition (SVD) and employs a scaling mechanism to constrain updates within distinct subspace scales. Together, these designs effectively reduce subspace interference and parameter drift. Extensive experiments on conventional benchmarks demonstrate that our method achieves state-of-the-art results.
\end{abstract}

\section{Introduction}
In real-world scenarios, deep learning models deployed in dynamic environments must continuously acquire new knowledge without forgetting previously learned concepts. However, stringent privacy constraints and limited storage often render historical data inaccessible. Under such strictly exemplar-free settings, sequential training inevitably biases model updates toward the current task, catastrophically disrupting prior representations~\cite{li2017learning}. This stability-plasticity dilemma is particularly acute in Class Incremental Learning (CIL)~\cite{rebuffi2017icarl}, which requires progressively expanding the recognition space without revisiting old data.

\begin{wrapfigure}[10]{r}{0.3\linewidth}
\vspace{-20pt}
\centering
\includegraphics[width=\linewidth]{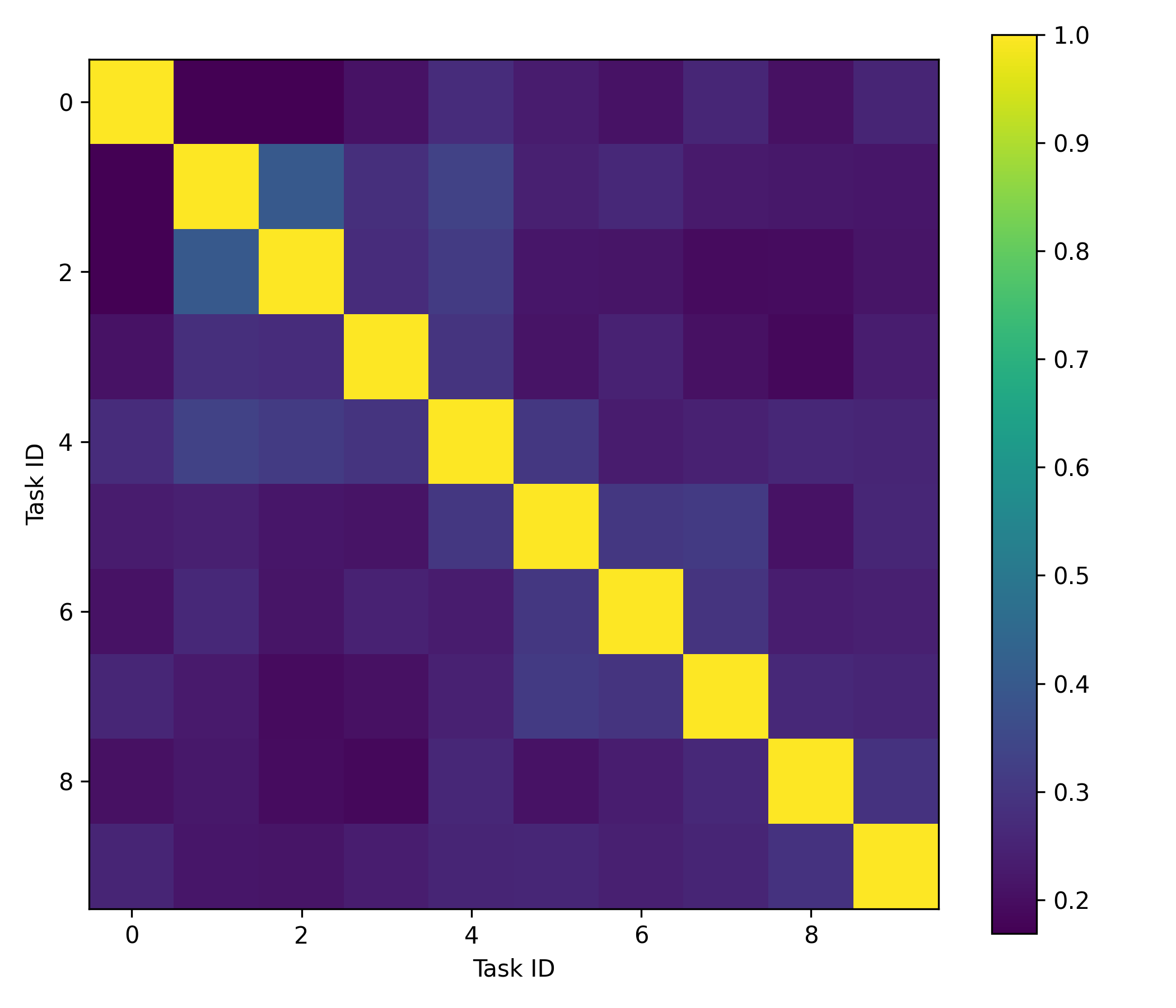}
\vspace{-20pt}
\caption{Subspace overlap on CIFAR-100 (B0 Inc10).}
\label{fig:subspace}
\end{wrapfigure}

Vision--language pre-trained models (VLMs) like CLIP~\cite{radford2021learning} have emerged as a promising paradigm for CIL. Thanks to their robust cross-modal alignment, continual adaptation requires updating only a marginal set of parameters. Consequently, recent methods typically freeze the backbone and insert lightweight tunable modules, such as prompts or adapters, to capture new task information~\cite{zhou2025learning,huang2024class}.

Despite their success, existing methods mainly constrain which parameters are updated, paying less attention to how these updates are organized in the parameter space. From a subspace perspective, updates induced by sequential tasks frequently collapse into low-rank subspaces with substantial overlap (as illustrated in \autoref{fig:subspace}). This structural overlap causes new-task updates to directly interfere with the optimal subspaces of previous tasks, leading to representation drift. We refer to this phenomenon as subspace interference, and regard it as an important source of catastrophic forgetting in CLIP-based CIL.

\begin{wrapfigure}[15]{r}{0.45\linewidth}
\vspace{-14pt}
\centering
\includegraphics[width=\linewidth]{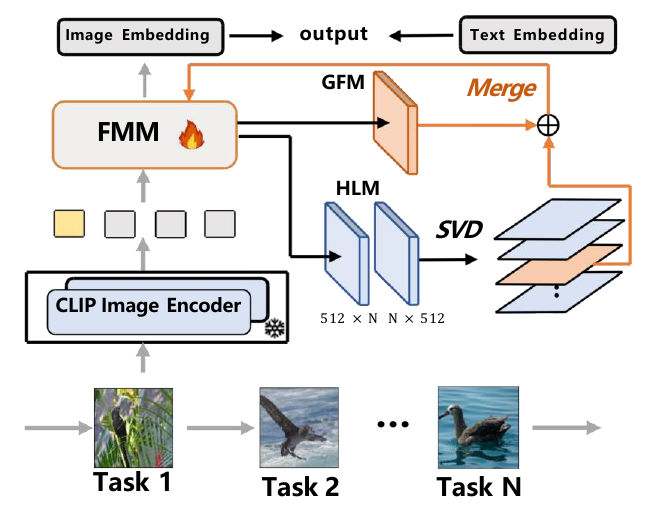}
\vspace{-20pt}
\caption{Overview of the proposed HDSD. FMM consists of GFM and HLM.}
\label{fig:small-framework}
\end{wrapfigure}

To tackle this issue, we propose a Hierarchical Dual-Subspace Decoupling (HDSD) framework, which explicitly governs cross-task parameter interactions. As shown in \autoref{fig:small-framework}, HDSD introduces a lightweight Feature Modulation Module (FMM) that factorizes the parameter space into shared and task-specific constituents through two complementary modules. First, the General Fusion Module (GFM) explicitly extracts and integrates generalized knowledge by measuring relative parameter variations across tasks, capturing highly stable components. Second, the Hierarchical Learning Module (HLM) isolates task-specific knowledge by performing singular value decomposition (SVD) on the parameter space. By assigning distinct tasks to separated, rank-one subspaces and regularizing their update magnitudes via a dynamic scaling strategy, HLM effectively suppresses cross-task interference. This overall design enables HDSD to balance general knowledge preservation and task-specific adaptation.

The main contributions of this paper are as follows:
\begin{itemize}
\item We propose the HDSD framework, which offers a novel perspective on CLIP-based CIL by explicitly decoupling the parameter space to simultaneously maximize knowledge sharing and task isolation.
\item We design the complementary GFM and HLM. GFM captures stable generalized knowledge, while HLM structurally prevents subspace interference via SVD-based parameter decomposition and scaled updates.
\item Extensive evaluations on standard benchmarks demonstrate that HDSD establishes a new state-of-the-art in rehearsal-free class incremental learning.
\end{itemize}

\section{Related Work}

\subsection{Class Incremental Learning}
Class Incremental Learning (CIL) aims to enable models to learn new classes in a continuously evolving environment without forgetting previously acquired knowledge, a challenge commonly referred to as the stability-plasticity dilemma~\cite{masana2022class}. 
To address this problem, existing methods can be broadly categorized into three groups. 
Regularization-based methods preserve previously learned knowledge by constraining parameter updates, typically through additional loss terms or optimization constraints that prevent significant changes to parameters important for past tasks, such as fine-tuning with regularization or joint training strategies~\cite{dhar2019learning,nokhwal2023rtra,sun2023regularizing,zhang2024regularization,wei2022incremental}. 
Replay-based methods alleviate forgetting by revisiting data from previous tasks, either by storing representative samples in a memory buffer or generating pseudo-data that can be replayed during training on new tasks~\cite{prabhu2020gdumb,wang2021triple,channappayya2023augmented,li2025pseudo,gu2022not}. 
Parameter isolation methods reduce interference by allocating dedicated parameters or model components to different tasks, thereby preventing updates for new tasks from overwriting previously learned representations, albeit at the cost of increased model complexity~\cite{wang2022coscl,yang2022continual,zhang2023continual,wang2023isolation,wang2025class}.

\subsection{CIL with Pre-trained Models}

With the success of large pre-trained models, recent CIL approaches increasingly leverage their strong generalization ability to mitigate catastrophic forgetting and improve data efficiency. 
These methods can be broadly divided into two categories based on their adaptation strategies. 
The first category performs full or partial fine-tuning of the pre-trained model, directly updating model parameters to adapt to new tasks, often with additional constraints or external data to stabilize training. 
For example, ZSCL~\cite{zheng2023preventing} leverages large-scale external data to regularize the adaptation process and preserve general representations. 
In contrast, the second category freezes the pre-trained backbone and introduces lightweight learnable components for task-specific adaptation, thereby preserving the general knowledge encoded in the backbone while confining task-specific updates to a smaller parameter space. 
Representative methods such as RAPF~\cite{huang2024class} achieve this by introducing adapter-based parameter fusion and generating hard sample pairs to improve robustness.

\subsection{CIL without Exemplar}

In many real-world applications, storing samples from previous tasks is often impractical due to privacy constraints and limited memory resources. 
To address this limitation, exemplar-free CIL methods aim to retain prior knowledge without relying on explicit data replay. 
Existing approaches can be broadly categorized into three directions. 
Distribution modeling methods approximate the feature distribution of previously learned classes, for example by leveraging Gaussian statistics to facilitate classification~\cite{hayes2020lifelong,tang2023prompt}. 
Prototype-based methods maintain compact class representations, such as feature centroids, to summarize historical knowledge while significantly reducing memory consumption~\cite{zhang2021prototype}. 
Generation-based methods synthesize pseudo samples or feature representations of past tasks, enabling implicit replay without storing raw data~\cite{gao2023unified,huang2024class}.

\section{Method}

\begin{figure}[t]  
\centering  
\includegraphics[width=\columnwidth]{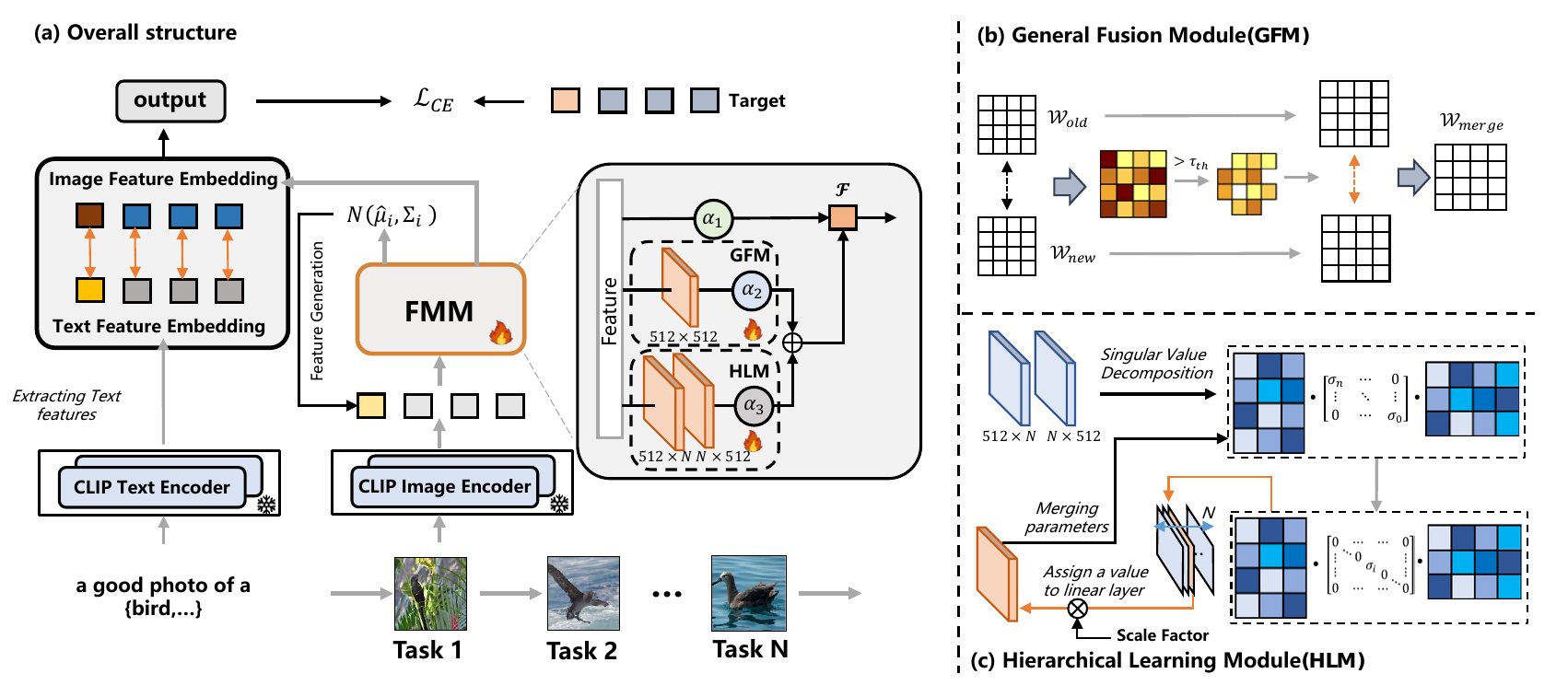}  
\caption{Summary of the  proposed approach. 
   (a) Overall structure features a Feature Modulation Module (FMM) consisting of two modules: GFM and HLM.      
   (b) Parameter Selection fusion method in GFM method: We calculate the relative change rate of GFM parameters to obtain the parameters related to general knowledge.      
   (c) Schematic of HLM method.       
   We segment the HLM parameters using the Singular Value Decomposition (SVD) method and assign segmented parameter subspaces to different incremental tasks to learn the corresponding specific knowledge.}  
\label{fig:overall-framework}  
\end{figure}

In this section, we formulate the CIL problem and detail the proposed Hierarchical Dual-Subspace Decoupling (HDSD) framework. The overall architecture is illustrated in \autoref{fig:overall-framework}. 

We build our method upon the pre-trained CLIP~\cite{radford2021learning} model, keeping its backbone parameters frozen throughout training. On top of it, we introduce a lightweight \textbf{Feature Modulation Module (FMM)} as the core learnable component for continual adaptation. 

The FMM consists of two complementary modules: the \textbf{General Fusion Module (GFM)} and the \textbf{Hierarchical Learning Module (HLM)}, as shown in \autoref{fig:overall-framework}(b) and (c). Specifically, GFM explicitly captures shared, stable knowledge across tasks, while HLM isolates task-specific knowledge through orthogonal subspace decomposition.

\subsection{Preliminaries}

Class Incremental Learning (CIL) aims to acquire knowledge from a sequence of evolving data streams, where each stream introduces a group of new classes. At each stage, the model has access only to the current data stream and must learn new knowledge without forgetting previously acquired information. We treat each data stream as an incremental task, denoted as $\{T_1, T_2, ..., T_N\}$, where $N$ is the total number of tasks. The $i$-th task $T_i$ contains data from $m$ classes, represented as $T_i = \{(x_b, y_b)\}$, where $x_b$ denotes an input image and $y_b \in \{1, ..., m\}$ is the corresponding class label. During training, the model is updated using only the data from the current task $T_i$. After each task, the model is evaluated on all classes encountered so far, requiring it to maintain performance on previous tasks while adapting to new ones.

\textbf{Feature Modulation Module.}
We first utilize the frozen CLIP image encoder to extract base features, denoted as $x$. These features are then fed into the FMM, which processes them through the GFM and HLM in parallel, accompanied by a residual connection. Let $x_G$ and $x_H$ represent the outputs of GFM and HLM, respectively. The final modulated feature is computed as:
\begin{align}
 x_{\mathrm{feature}} = \alpha_1 x + \alpha_2 x_G + \alpha_3 x_H
 \label{fenjie}
\end{align}
where $\alpha_1$, $\alpha_2$, and $\alpha_3$ are weighting coefficients, set to $(1, 0.5, 0.5)$ by default.

\textbf{Feature Generation.}
To circumvent the need for storing raw historical images (exemplar-free), we leverage the robust representation space of the pre-trained model to approximate the feature distribution of each class. For class $b$, we compute the centroid $\mu_b$ and covariance matrix $\Sigma_b$ from the CLIP-extracted features. By drawing samples from the multivariate Gaussian $\mathcal{N}(\mu_b, \Sigma_b)$, we synthesize pseudo-features for implicit memory replay during subsequent tasks.

\subsection{General Fusion Module}

The General Fusion Module (GFM) is parameterized by a linear layer $\theta$ and is designed to identify and preserve features that generalize across tasks.

\textbf{Sparse Regularization.}
We introduce an $\ell_1$-norm sparsity penalty, $\mathcal{L}_{\text{sparse}}$, to reduce model complexity and encourage compact representations, thereby enhancing GFM's ability to extract generalized features:
\begin{align}
 \mathcal{L}_{\text{sparse}} = \beta \sum_{p_i \in \theta} \| p_i \|_1,
\end{align}
where $p_i$ denotes the $i$-th element of parameter tensor $p$, and $\beta$ is a weighting hyperparameter set to $0.0005$ by default. 

\textbf{Parameter Selection and Fusion.}
To enhance the retention of general knowledge, we perform parameter selection and fusion based on their changes across consecutive tasks. 
For the incremental task $T_i$, we denote $\theta_{\text{old}}$ as the GFM parameters obtained after training on the previous task $T_{i-1}$, and $\theta_{\text{new}}$ as the updated parameters after training on the current task $T_i$. 
We then compute the relative change rate:
\begin{align}
 \Gamma = \frac{|\theta_{\text{new}} - \theta_{\text{old}}|}{\max |\theta_{\text{new}} - \theta_{\text{old}}|} + c
\end{align}

where $c \in [0.5, 0.7]$ is a small constant for numerical stabilization.
To distinguish between general and task-specific parameters, we introduce an adaptive threshold $\tau$ based on the distribution of $\Gamma$. 
Specifically, let $\{\Gamma_j\}$ denote all elements of $\Gamma$ flattened into a vector. 
We define $\tau$ as the $q$-th percentile of $\{\Gamma_j\}$:
\begin{align}
\tau = \mathrm{Quantile}(\{\Gamma_j\}, q)
\end{align}
where $q = 0.9$ in our implementation. Intuitively, parameters exhibiting significant variations ($\Gamma_j > \tau$) are highly task-specific, whereas parameters with minimal changes ($\Gamma_j \le \tau$) represent stable, generalized knowledge.

Consequently, for each parameter index $j$, we perform a weighted fusion:
\begin{align}
M_j =
\begin{cases}
(1 - \min(1, \Gamma_j)) \cdot \theta_{\text{old}, j} + \min(1, \Gamma_j) \cdot \theta_{\text{new}, j}, & \Gamma_j \le \tau \\
\theta_{\text{old}, j}, & \Gamma_j > \tau
\end{cases}
\end{align}

This element-wise strategy preserves historically stable parameter directions while affording sufficient plasticity for novel task adaptation.

\subsection{Hierarchical Learning Module}
The Hierarchical Learning Module (HLM) is designed to explicitly map task-specific parameters into isolated subspaces, mitigating cross-task interference via structured SVD decomposition and controlled updates.

\textbf{Parameter Subspace Decomposition.}
HLM consists of two linear layers with an intermediate bottleneck dimension set to $N$ (the total number of incremental tasks). Let $\mathcal{W}$ denote the target weight matrix. We perform Singular Value Decomposition (SVD):
\begin{align}
\mathcal{W} = U \Sigma V^T
\end{align}
where $\Sigma = \mathrm{diag}(\sigma_1, \sigma_2, ..., \sigma_N)$ contains $N$ singular values. 
This decomposition allows us to express $\mathcal{W}$ as a sum of $N$ rank-1 components:
\begin{align}
\mathcal{W} = \sum_{i=1}^{N} \mathcal{W}_{\sigma_i}, \quad \text{where } \mathcal{W}_{\sigma_i} = U \sigma_i V^T
\end{align}
Each $\mathcal{W}_{\sigma_i}$ corresponds to a \textbf{task-specific parameter subspace}.

\begin{wrapfigure}[14]{r}{0.4\linewidth}
\vspace{-14pt}
\centering
\includegraphics[width=\linewidth]{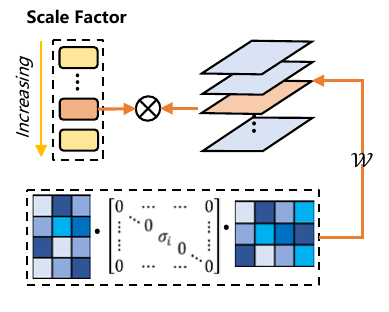}
\vspace{-20pt}
\caption{The training phase of our proposed hierarchical approach in HLM.}
\label{fig:HLM-train}
\end{wrapfigure}

\textbf{Hierarchical Training.}
As illustrated in \autoref{fig:HLM-train}, during task $T_i$, optimization is strictly confined to its corresponding subspace $\mathcal{W}_{\sigma_i}$. Specifically, $\mathcal{W}_{\sigma_i}$ is scaled by a factor $s_i$ and assigned to a temporary linear layer. Post-training, this updated subspace is rescaled and merged back. This enforces structural task isolation.

\textbf{Stability and Scaling Mechanism.}
Although HLM decomposes the parameter matrix into multiple singular components, two stability issues need to be considered during sequential training.

\textit{Problem 1: Stability of Singular Component Ordering.}

A significant change in a singular value may alter the order of singular components after SVD, making it difficult to retrieve the component associated with a specific task.
To avoid this issue, HLM updates only the singular value corresponding to the current task while keeping the others inactive.
For task $i$, only $\mathcal{W}_{\sigma_i}$ is optimized.
Since the update magnitude is controlled by the learning rate and the scaling factor, the variation of $\sigma_i$ remains small in practice.
Therefore, the order of singular components is stable, and each task-specific component can be retrieved according to its original training order.

\textit{Problem 2: Drift in Shared Singular Bases ($U$ and $V^T$).}





Despite decoupled singular values, all task components share the orthogonal bases $U$ and $V^T$. Unrestricted updates to task $i$ could implicitly rotate these bases, corrupting previously learned subspaces. To decouple this interaction, we introduce an exponential task-dependent scaling factor:
\begin{align}
s_i = 10^{-i},
\end{align}
where $i$ is the current task index. Prior to optimization, the component is scaled: $\widetilde{\mathcal{W}}_{\sigma_i} = s_i \mathcal{W}_{\sigma_i}$. The gradient update is then applied:
\begin{align}
\theta_{t+1} = \theta_t - \eta \nabla J(\theta_t),
\end{align}

Because sequential tasks are separated by orders of magnitude, $s_i$ severely constrains the effective update momentum exerted on the shared bases. Following optimization, the component is inverse-scaled ($\mathcal{W}_{\sigma_i}^{\text{new}} = \widetilde{\mathcal{W}}_{\sigma_i}^{\text{new}} / s_i$) and aggregated:
\begin{align}
\mathcal{W}^{\text{new}} = \sum_{j=1}^{N} \mathcal{W}_{\sigma_j}^{\text{new}}.
\end{align}
A comprehensive theoretical discussion on how this scaling averts basis drift is provided in \autoref{sec:supp-scaling}.

\textbf{Hierarchical Loss.}
To maximize task-specific discriminability within these subspaces, we apply a hierarchical centroid-based loss:
\begin{align}
\mathcal{L}_{H} = \frac{\xi}{2n} \sum_{i=1}^{n} \frac{1}{C_{y_i}} \left\| f(x_i) - c_{y_i} \right\|^2
\end{align}
where $\xi=0.2$, $n$ is the batch size, $C_{y_i}$ is the number of samples in class $y_i$, and $c_{y_i}$ is the class prototype centroid.

\textbf{Overall Objective.}
The final training objective is:
\begin{align}
\mathcal{L} = \lambda_1 \mathcal{L}_{\text{sparse}} + \lambda_2 \mathcal{L}_{H} + \lambda_3 \mathcal{L}_{\text{ce}}
\end{align}

where $\mathcal{L}_{\text{ce}}$ is the standard cross-entropy loss, and $\lambda_1, \lambda_2, \lambda_3$ are weighting coefficients, set to (0.01, 0.1, 1) by default.

\begin{wrapfigure}[16]{r}{0.4\linewidth}
\vspace{-14pt}
\centering
\includegraphics[width=\linewidth]{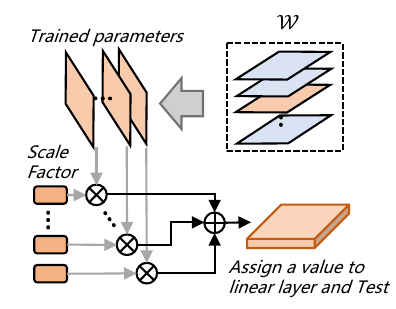}
\vspace{-20pt}
\caption{The test phase of our proposed hierarchical approach in HLM.}
\label{fig:HLM-test}
\end{wrapfigure}
\textbf{Hierarchical Testing.}
As illustrated in \autoref{fig:HLM-test}, the inference stage differs fundamentally from training in that it aggregates information from all previously learned tasks rather than operating on a single task-specific component. 
While training updates only one subspace at a time, inference reconstructs the full parameter matrix by hierarchically integrating all learned subspaces.

Specifically, after training up to task $T_i$, we collect the corresponding singular values $\{\sigma_1, \sigma_2, \dots, \sigma_i\}$ along with their associated scaling factors $\{s_1, s_2, \dots, s_i\}$, and construct a diagonal matrix:
\begin{align}
A = \mathrm{diag}(s_1 \sigma_1, s_2 \sigma_2, ..., s_i \sigma_i)
\end{align}

The parameter matrix is then reconstructed as:
\begin{align}
\mathcal{W}_{\text{test}} = U A V^T
\end{align}

Notably, the scaling factors introduced during training are preserved at inference time, resulting in a hierarchical weighting over task-specific components.

\section{Experiment}
\subsection{Datasets and Metrics}
We conduct experiments on three datasets: CIFAR-100~\cite{krizhevsky2009learning}, ImageNet-100, and ImageNet-R~\cite{hendrycks2021many}.  
The CIFAR-100 dataset consists of 100 categories, each containing 600 color images with a resolution of $32\times 32$ pixels.   
In this dataset, 500 images are designated for the training set, while 100 images are reserved for the test set.   
The ImageNet-100 dataset, a subset of the larger ImageNet-1k dataset~\cite{krizhevsky2012imagenet}, comprises 100 categories. 
Each category contains 1,000 training images and 300 test images, resulting in a total of 100,000 training images and 30,000 test images across all categories.
Meanwhile, the ImageNet-R dataset features images in various styles, including art, cartoons, graffiti,embroidery, and video games, among others, and consists of 200 classes from the ImageNet dataset. Following the methodology proposed in \cite{huang2024class}, we split this dataset into training and test sets.  

\textbf{Evaluation Metrics.} We report two standard metrics: \textbf{Last}, the average accuracy over all encountered categories after the final incremental task, and \textbf{Avg}, the average of the seen-category accuracies computed after each incremental step.

\begin{table}[t]
\centering
\footnotesize
\setlength{\tabcolsep}{4.5pt}
\renewcommand{\arraystretch}{1.0}

\caption{Experimental results for continual learning on the ImageNet-100 dataset.
The B denotes the number of base classes and Inc denotes the number of incremental classes.}
\label{tab:imagenet100}

\begin{tabular}{l c cc cc cc cc cc}
\toprule
\multirow{2}{*}{Method} & \multirow{2}{*}{Ref.}
& \multicolumn{2}{c}{B0 Inc5}
& \multicolumn{2}{c}{B0 Inc10}
& \multicolumn{2}{c}{B0 Inc20}
& \multicolumn{2}{c}{B50 Inc5}
& \multicolumn{2}{c}{B50 Inc10} \\
\cmidrule(lr){3-4} \cmidrule(lr){5-6} \cmidrule(lr){7-8} \cmidrule(lr){9-10} \cmidrule(lr){11-12}
& & Avg & Last & Avg & Last & Avg & Last & Avg & Last & Avg & Last \\
\midrule

PROOF~\cite{zhou2025learning} & TPAMI'25 
& 86.92 & 75.52 & 84.71 & 72.48 & 81.92 & 68.56 & 84.16 & 74.44 & 82.78 & 71.04 \\
\midrule

L2P++~\cite{wang2022learning} & CVPR'22 
& 75.43 & 62.10 & 80.51 & 67.22 & 84.12 & 73.70 & 62.00 & 22.15 & 74.11 & 49.46 \\

DualPrompt~\cite{wang2022dualprompt} & ECCV'22 
& 75.40 & 61.10 & 80.65 & 67.38 & 84.65 & 74.24 & 62.10 & 22.36 & 74.20 & 49.78 \\

CODA~\cite{smith2023coda} & CVPR'23 
& 51.64 & 24.94 & 64.13 & 34.76 & 69.78 & 43.96 & 57.33 & 19.95 & 65.14 & 28.80 \\

ADAM-Adapter~\cite{zhou2025revisiting} & IJCV'25 
& 85.78 & 75.52 & 85.84 & 76.40 & 85.85 & 77.08 & \underline{84.90} & 78.58 & 84.60 & 78.58 \\

ENGINE~\cite{zhou2025external} & ICCV'25  & 86.34 & 76.82  & 85.65  & 76.76 & 85.08 & 76.60 & 80.83 & 76.82 & 80.68 & 76.54 \\

BOFA~\cite{li2026bofa} & AAAI'26  & 83.99 & 73.80 & 84.78 & 75.36 & 83.61 & 74.98 & 81.50  & 76.08 & 81.23 & 76.16 \\

RAPF~\cite{huang2024class} & ECCV'24 
& \underline{87.55} & \underline{79.40} & \underline{87.34} & \underline{80.60} 
& \underline{87.28} & \underline{80.00} & 84.74$^*$ & \underline{80.44} & \underline{84.96$^*$} & \underline{80.66} \\

\midrule
Ours & -- 
& \textbf{89.49} & \textbf{80.82} & \textbf{89.14} & \textbf{81.44} 
& \textbf{88.67} & \textbf{81.10} & \textbf{85.14} & \textbf{81.24} 
& \textbf{85.22} & \textbf{81.72} \\

\bottomrule
\end{tabular}
\end{table}
\begin{table}[ht]
\centering
\footnotesize
\setlength{\tabcolsep}{4.5pt}
\renewcommand{\arraystretch}{1.0}

\caption{Experimental results for continual learning on the CIFAR-100 dataset.
The B denotes the number of base classes and Inc denotes the number of incremental classes.}
\label{tab:cifar100}

\begin{tabular}{l c cc cc cc cc cc}
\toprule
\multirow{2}{*}{Method} & \multirow{2}{*}{Ref.}
& \multicolumn{2}{c}{B0 Inc5}
& \multicolumn{2}{c}{B0 Inc10}
& \multicolumn{2}{c}{B0 Inc20}
& \multicolumn{2}{c}{B50 Inc5}
& \multicolumn{2}{c}{B50 Inc10} \\
\cmidrule(lr){3-4} \cmidrule(lr){5-6} \cmidrule(lr){7-8} \cmidrule(lr){9-10} \cmidrule(lr){11-12}
& & Avg & Last & Avg & Last & Avg & Last & Avg & Last & Avg & Last \\
\midrule
PROOF~\cite{zhou2025learning}       & TPAMI'25 & 85.12 & 76.13 & 84.88 & 76.29 & 84.11 & 76.86 & 83.22 & 76.25 & 83.17 & 76.50 \\
\midrule

L2P++~\cite{wang2022learning}       & CVPR'22  & 79.18 & 68.67 & 81.90 & 73.08 & 84.39 & 77.37 & 58.57 & 18.04 & 76.51 & 48.52 \\

DualPrompt~\cite{wang2022dualprompt}  & ECCV'22  & 79.74 & 69.91 & 81.45 & 72.51 & 85.19 & 77.47 & 58.55 & 15.26 & 72.00 & 45.05 \\

CODA~\cite{smith2023coda}       & CVPR'23  & 69.78 & 41.98 & 76.98 & 62.25 & 78.65 & 65.29 & 58.45 & 15.99 & 67.88 & 28.77 \\

ADAM-Adapter~\cite{zhou2025revisiting} & IJCV'25  & 70.18 & 58.12 & 75.76 & 65.50 & 77.28 & 67.89 & \underline{83.38} & 76.94 & 83.21 & 76.94 \\

ENGINE~\cite{zhou2025external} & ICCV'25  & 86.42 & 78.58  & 86.21  & 78.64 & \underline{85.72} & 78.82 & 81.99 & 78.51 & 82.28 & 78.49 \\

BOFA~\cite{li2026bofa} & AAAI'26  & 86.41 & 77.86 & 85.72 & 78.28 & 85.31 & 78.89 & 82.92  & 78.47 & 82.99 & 78.70\\

RAPF~\cite{huang2024class}  & ECCV'24  & \underline{86.67} & \underline{78.84} & \underline{85.89} & \underline{79.17} & 85.46 & \underline{79.97} & 82.99$^*$ & \underline{79.00} & \underline{83.53$^*$} & \underline{80.33} \\

\midrule
Ours        & --       & \textbf{87.53} & \textbf{80.01} & \textbf{87.10} & \textbf{80.86} & \textbf{86.53} & \textbf{81.24} & \textbf{83.41} & \textbf{80.18} & \textbf{83.72} & \textbf{80.96} \\
\bottomrule
\end{tabular}
\end{table}

\subsection{Implementation Detail}
We adopt the ViT-B/16 variant of CLIP~\cite{radford2021learning} as our frozen backbone. The model is optimized using the Adam optimizer. For ImageNet-100 and ImageNet-R, we train the model for 15 epochs per task, initializing the learning rates for GFM and HLM at 0.001 and 0.01, respectively. For CIFAR-100, the initial training spans 25 epochs using the same learning rates, and we progressively increase the training duration by 2 epochs for each subsequent incremental task. A MultiStepLR scheduler is employed to decay the learning rate by a factor of 0.1 at epochs 4 and 10. To simulate memory replay without storing real images, we generate approximately 2,000 pseudo-feature samples per epoch, matching the typical buffer size of traditional replay methods.

\begin{table}[t]
\centering
\footnotesize
\setlength{\tabcolsep}{5pt}
\renewcommand{\arraystretch}{1.0}

\caption{Experimental results for continual learning on the ImageNet-R dataset.
The B denotes the number of base classes and Inc denotes the number of incremental classes.}
\label{tab:imagenet-r}

\begin{tabular}{l c cc cc cc cc cc}
\toprule
\multirow{2}{*}{Method} & \multirow{2}{*}{Ref.}
& \multicolumn{2}{c}{B0 Inc10}
& \multicolumn{2}{c}{B0 Inc20}
& \multicolumn{2}{c}{B0 Inc40}
& \multicolumn{2}{c}{B100 Inc10}
& \multicolumn{2}{c}{B100 Inc20} \\
\cmidrule(lr){3-4} \cmidrule(lr){5-6} \cmidrule(lr){7-8} \cmidrule(lr){9-10} \cmidrule(lr){11-12}
& & Avg & Last & Avg & Last & Avg & Last & Avg & Last & Avg & Last \\
\midrule

PROOF~\cite{zhou2025learning} & TPAMI'25
& 82.69 & 77.25 & 82.83 & 77.05 & 82.63 & 77.12 & 81.61 & 77.10 & 81.78 & 77.17 \\
\midrule

L2P++~\cite{wang2022learning} & CVPR'22
& 76.87 & 68.78 & 81.67 & 75.98 & 82.81 & 77.87 & 56.17 & 17.90 & 67.73 & 43.28 \\

DualPrompt~\cite{wang2022dualprompt} & ECCV'22
& 77.07 & 69.41 & 82.01 & 75.77 & 83.77 & 78.64 & 57.37 & 19.18 & 69.18 & 45.37 \\

CODA~\cite{smith2023coda} & CVPR'23
& 75.23 & 64.53 & 78.00 & 67.52 & 78.80 & 71.27 & 56.62 & 17.64 & 65.62 & 35.06 \\

ADAM-Adapter~\cite{zhou2025revisiting} & IJCV'25
& 76.71 & 68.75 & 78.65 & 71.35 & 79.87 & 73.02 & 79.87 & 75.37 & 79.75 & 75.37 \\

ENGINE~\cite{zhou2025external} & ICCV'25  & 86.04 & 80.12 & \underline{86.10} & \underline{80.72} & \underline{85.88} & \underline{81.13} & 82.95  & 80.35 & 83.60 & 81.00 \\

BOFA~\cite{li2026bofa} & AAAI'26  & 86.10 & 79.97 & 85.42 & 79.78 & 84.35 & 79.92 & 81.71 & 79.80 & 81.73 & 79.82 \\

RAPF~\cite{huang2024class} & ECCV'24
& \underline{86.21} & \underline{80.23} & 85.25 & 80.58 & 84.52 & 80.91 
& \underline{83.43} & \underline{81.37} & \underline{83.77} & \underline{81.45} \\

\midrule
Ours & --
& \textbf{88.15} & \textbf{82.11} & \textbf{88.24} & \textbf{82.46} 
& \textbf{87.57} & \textbf{82.92} & \textbf{84.77} & \textbf{82.57} 
& \textbf{84.55} & \textbf{82.13} \\

\bottomrule
\end{tabular}
\end{table}

\subsection{Experimental Results}

We benchmark HDSD against several state-of-the-art CIL methods: PROOF~\cite{zhou2025learning}, L2P++~\cite{wang2022learning}, DualPrompt~\cite{wang2022dualprompt}, CODA~\cite{smith2023coda}, ADAM-Adapter~\cite{zhou2025revisiting}, ENGINE~\cite{zhou2025external}, BOFA~\cite{li2026bofa}, and RAPF~\cite{huang2024class}. The comparative results across the three datasets are presented in \autoref{tab:imagenet100}, \autoref{tab:cifar100}, and \autoref{tab:imagenet-r}. Notably, while PROOF relies on exemplars, all other baselines are strictly exemplar-free. 

As shown in \autoref{tab:imagenet100} and \autoref{tab:cifar100}, the results we reproduced for RAPF differ slightly from those reported in the original paper. The settings with relatively large discrepancies (around 2\% compared with the reported results) are marked by $^*$. These differences are mainly observed in the B50 Inc5 and B50 Inc10 settings on both CIFAR-100 and ImageNet-100.

\textbf{Performance on ImageNet-100.} As shown in \autoref{tab:imagenet100}, our method achieves the best performance in all settings on the ImageNet-100 dataset. 
Under the B0 setting, our method improves the Avg and Last accuracy over RAPF by 1.71\% and 1.12\% on average, respectively. 
Under the B50 setting, the Avg and Last accuracy are further improved by 0.33\% and 0.93\% on average, respectively. 
These results demonstrate that our method yields consistent gains across all ImageNet-100 settings.

\textbf{Performance on CIFAR-100.} As shown in \autoref{tab:cifar100}, our method achieves the best performance across all settings on the CIFAR-100 dataset. 
Under the B0 setting, our method improves the Avg and Last accuracy over RAPF by 1.05\% and 1.38\% on average, respectively. 
Under the B50 setting, the Avg and Last accuracy are improved by 0.31\% and 0.91\% on average, respectively. 
These results indicate that our method consistently enhances both performance and stability during incremental learning.

\textbf{Performance on ImageNet-R.} As shown in \autoref{tab:imagenet-r}, our method achieves the best performance across all settings on the ImageNet-R dataset. 
Under the B0 setting, our method improves the Avg and Last accuracy over RAPF by 2.66\% and 1.92\% on average, respectively. 
Under the B100 setting, the Avg and Last accuracy are improved by 1.06\% and 0.94\% on average, respectively. 
These results demonstrate the effectiveness of our method under both standard and more challenging settings.

\begin{figure}[t]
    \centering
    \includegraphics[width=0.48\linewidth]{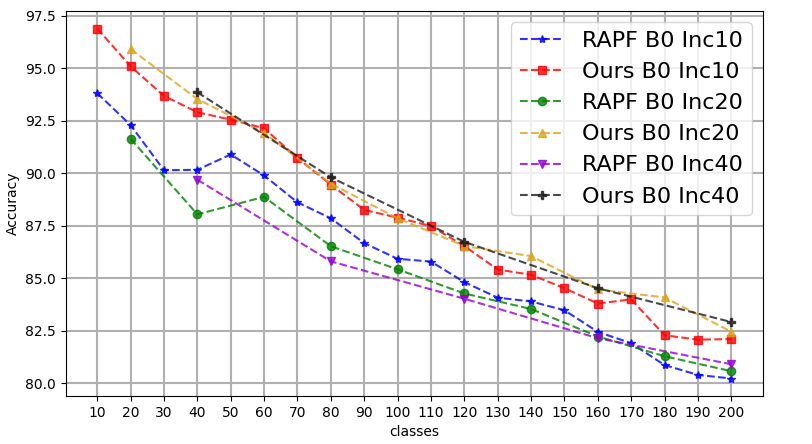}
    \includegraphics[width=0.48\linewidth]{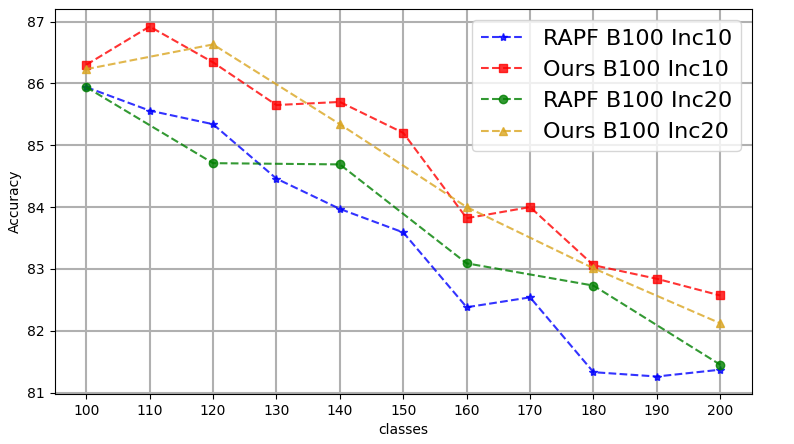}
    \caption{
    Accuracy curves of RAPF and our method on ImageNet-R.
    The left figure shows the results under the B0 setting, and the right figure shows the results under the B100 setting.
    }
    \label{fig:imagenet-r-curve}
\end{figure}

\textbf{Learning Dynamics.} To further analyze the learning dynamics, we visualize the accuracy curves of RAPF and our method on ImageNet-R in \autoref{fig:imagenet-r-curve}.
The corresponding results on ImageNet-100 and CIFAR-100 are provided in \autoref{sec:supp-learning-curves}.
Compared with RAPF, our method maintains higher accuracy during the whole incremental process under both B0 and B100 settings.
In particular, the advantage is more pronounced in the B0 setting, where no base classes are provided and the model needs to learn all classes sequentially from scratch.
As the number of learned classes increases, RAPF suffers from a more noticeable performance drop, while our method shows a smoother degradation trend.
This indicates that the proposed subspace modeling strategy can better preserve previously learned knowledge and reduce accumulated forgetting.
In the B100 setting, although the performance gap becomes smaller due to the strong initialization provided by the base session, our method still consistently outperforms RAPF across different incremental steps.

\textbf{Key Observations.} From the above results, we observe two notable trends. First, the performance gains on CIFAR-100 are relatively smaller than those on ImageNet-100 and ImageNet-R. This may be because CIFAR-100 contains low-resolution images with subtle inter-class differences, making the model more sensitive to parameter variations. While constraining updates within decomposed subspaces improves stability, it may also reduce the flexibility of shared parameters such as $U$ and $V^T$, leading to less pronounced improvements on CIFAR-100.
Second, our method brings larger gains in the B0 setting, while the improvements become smaller when a non-zero base session is used. We attribute this to the fixed subspace allocation strategy, where the base session is assigned the same capacity as each subsequent incremental task. Although this allocation is sufficient for small increments, it can be restrictive for larger base sessions, such as 50 or 100 base classes, which require greater model capacity. Overall, our method achieves consistently strong performance across datasets, demonstrating a favorable balance between plasticity and stability.

\subsection{Threshold Selection}

\begin{wrapfigure}{r}{0.4\linewidth}
  \centering
  \vspace{-10pt}
  \includegraphics[width=\linewidth]{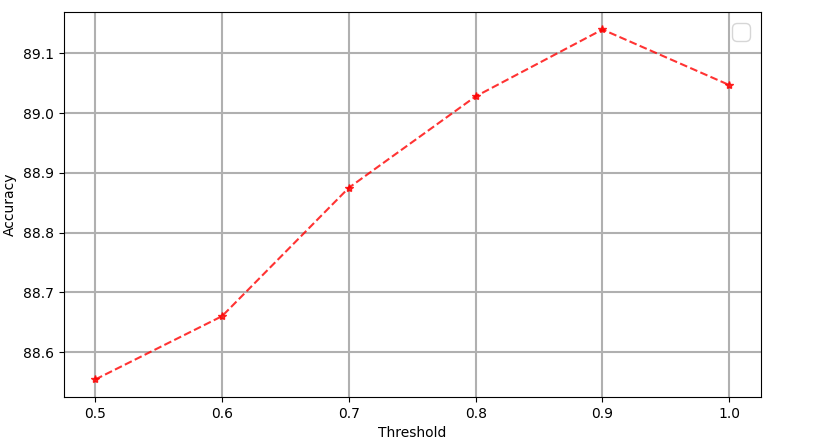}
  \caption{Performance under different threshold values on ImageNet-100 (B0 Inc10).}
  \label{zonghe1}
  \vspace{-10pt}
\end{wrapfigure}

The threshold $\tau$ used in the General Fusion Module (GFM) is defined based on the distribution of the relative parameter change $\Gamma$. 
To determine an appropriate value, we conduct experiments on the ImageNet-100 dataset with 0 base classes and an increment size of 10, resulting in a total of 10 tasks. 
We vary the percentile parameter $q$ in the range of $[0.5, 1.0]$, which controls the threshold $\tau = \mathrm{Quantile}(\{\Gamma_j\}, q)$.

As shown in \autoref{zonghe1}, performance consistently improves as $q$ increases from $0.5$ to $0.9$, suggesting that aggressively filtering out unstable parameters helps preserve the shared representation. 
However, further increasing $q$ beyond $0.9$ leads to a slight performance drop, indicating that overly restrictive selection may hinder the model’s ability to adapt to new tasks.

The best performance is achieved at $q = 0.9$, which corresponds to selecting the top $10\%$ of parameters with the largest relative changes as task-specific components. Based on this observation, we fix $q = 0.9$ for all experiments unless otherwise specified.

\subsection{Ablation Experiment}

\begin{wraptable}{r}{0.48\linewidth}
 
  \centering
  \caption{Ablation study on ImageNet-R under the B0 Inc20 setting.}
  \footnotesize
  \setlength{\tabcolsep}{2pt}
  \renewcommand{\arraystretch}{1.05}
  
  \begin{tabular}{l c c c c c}
    \toprule
    Method & GFM & HLM & $\mathcal{L}_{\text{sparse}}$ & $\mathcal{L}_{H}$ & Avg \\
    \midrule
    Baseline &  &  &  &  & 85.20 \\
    + HLM + $\mathcal{L}_{H}$ &  & \checkmark &  & \checkmark & 86.25 \\
    + GFM + $\mathcal{L}_{\text{sparse}}$ & \checkmark &  & \checkmark &  & 86.36 \\
    + GFM + HLM + $\mathcal{L}_{H}$ & \checkmark & \checkmark &  & \checkmark & 86.33 \\
    + GFM + HLM + $\mathcal{L}_{\text{sparse}}$ & \checkmark & \checkmark & \checkmark &  & 86.02 \\
    \midrule
    Ours & \checkmark & \checkmark & \checkmark & \checkmark & \textbf{88.24} \\
    \bottomrule
  \end{tabular}

  \label{table3}
  \vspace{-10pt}
\end{wraptable}

We conduct ablation experiments on the ImageNet-R dataset with 0 base classes and an increment size of 20. The results are reported in \autoref{table3}. From the table, we observe that the General Fusion Module (GFM) contributes the largest performance gain, demonstrating its effectiveness in capturing stable and shared knowledge across tasks. The Hierarchical Learning Module (HLM) also provides clear improvements, while the sparse regularization term $\mathcal{L}_{\text{sparse}}$ further enhances performance by encouraging compact representations.

In contrast, the contribution of $\mathcal{L}_{H}$ is relatively limited when used without the complete model design. When all components are combined, our full model achieves the best performance, outperforming all partial variants by a clear margin. These results verify that the proposed modules are complementary and jointly improve both representation quality and incremental stability.

\section{Conclusion}

In this paper, we propose HDSD, a Hierarchical Dual-Subspace Decoupling framework for CLIP-based class incremental learning. HDSD addresses catastrophic forgetting by explicitly modeling inter-task parameter interactions in the subspace space. Through a lightweight Feature Modulation Module, GFM preserves general knowledge by fusing stable shared components, while HLM learns task-specific knowledge through separated subspaces with controlled updates. Experiments on CIFAR-100, ImageNet-100, and ImageNet-R show that HDSD consistently improves incremental learning performance over existing methods, demonstrating its effectiveness in reducing cross-task interference and balancing stability and plasticity.

\section*{Impact Statement}

This work proposes HDSD, a Hierarchical Dual-Subspace Decoupling framework for CLIP-based class incremental learning. All training and evaluation are conducted on publicly available benchmark datasets under standard experimental settings that are consistent with prior continual learning studies. The proposed approach focuses on algorithmic improvements and does not involve new data collection or real-world deployment, and therefore does not introduce additional privacy, security, or ethical concerns beyond those commonly considered in existing continual learning research. However, our method still has a limitation: HLM adopts a fixed subspace allocation strategy, which assigns the same capacity to the base session and each subsequent incremental task. This may restrict the representation capacity for non-zero base-session settings, leading to less significant performance gains when the number of base classes is large.

\bibliographystyle{unsrtnat}
\bibliography{nips}

\clearpage
\appendix

\section{Additional Learning Curves}
\label{sec:supp-learning-curves}

\begin{figure}[ht]
\centering
\begin{subfigure}{0.48\linewidth}
   \centering
    \includegraphics[width=\linewidth]{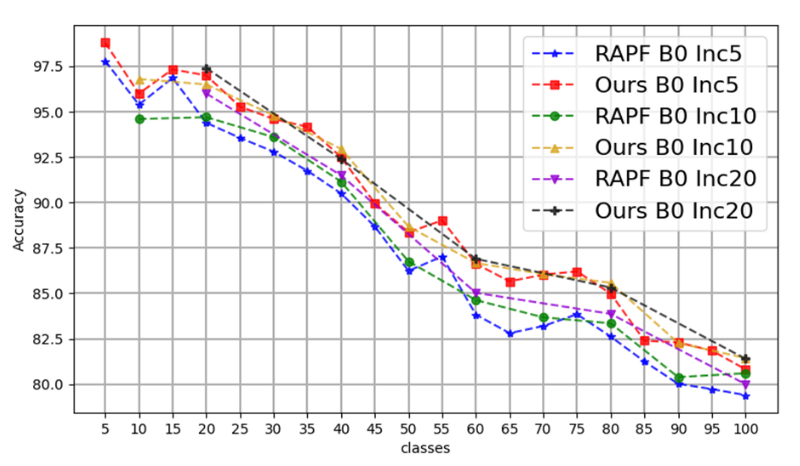}
    \caption{Comparison of the results of RAPF and our method on Imagenet-100 dataset. The result when there are 0 base classes.}
    \label{fig:supp-imagenet100-b0}
\end{subfigure}
\hfill
\begin{subfigure}{0.48\linewidth}
    \centering
    \includegraphics[width=\linewidth]{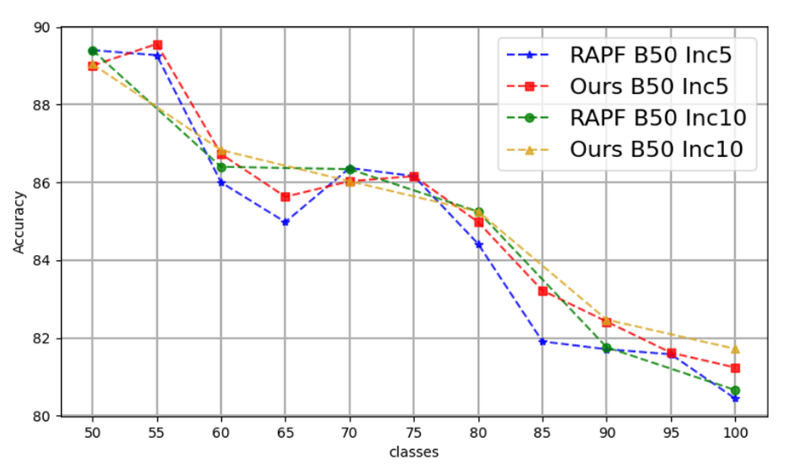}
    \caption{Comparison of the results of RAPF and our method on Imagenet-100 dataset. The result when there are 50 base classes.}
    \label{fig:supp-imagenet100-b50}
 
\end{subfigure}

\vspace{4pt}

\begin{subfigure}{0.48\linewidth}
    \centering
    \includegraphics[width=\linewidth]{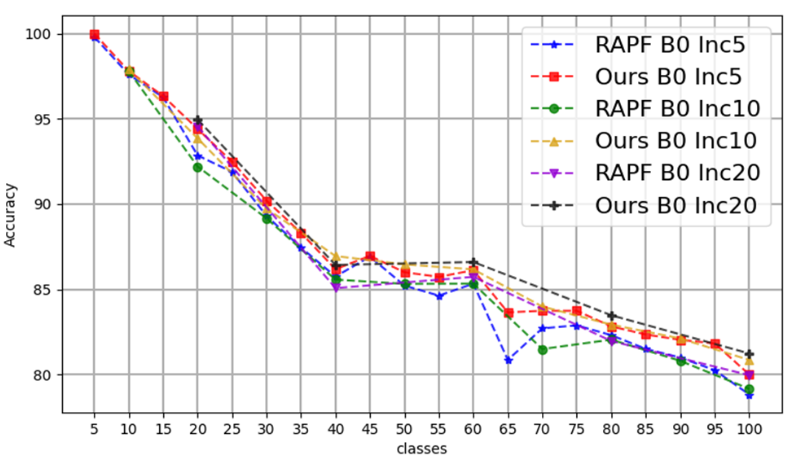}
    \caption{Comparison of the results of RAPF and our method on CIFAR-100 dataset. The result when there are 0 base classes.}
    \label{fig:supp-cifar100-b0}

\end{subfigure}
\hfill
\begin{subfigure}{0.48\linewidth}
    \centering
    \includegraphics[width=\linewidth]{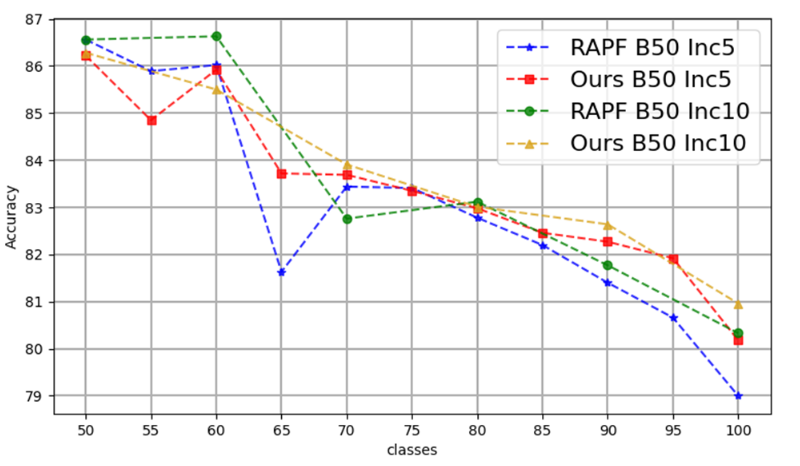}
    \caption{Comparison of the results of RAPF and our method on CIFAR-100 dataset. The result when there are 50 base classes.}
    \label{fig:supp-cifar100-b50}

\end{subfigure}

\caption{Additional accuracy curves on ImageNet-100 and CIFAR-100 under different base-session settings.}
\label{fig:supp-additional-learning-curves}
\end{figure}

The learning curves on ImageNet-100 and CIFAR-100 are shown in \autoref{fig:supp-imagenet100-b0}, \autoref{fig:supp-imagenet100-b50},  \autoref{fig:supp-cifar100-b0}
 and \autoref{fig:supp-cifar100-b50}.

\section{Discussion on the Scaling Mechanism}
\label{sec:supp-scaling}

In this section, we provide an additional discussion on the scaling mechanism used in HLM. 
The goal of the scaling factor is to reduce the influence of current-task updates on shared parameters and previously learned subspaces.

Consider the optimization of the task-specific component associated with task $T_i$. 
Let $W_{\sigma_i}$ denote the corresponding subspace component, and let $s_i$ be its scaling factor. 
Before optimization, HLM scales this component as
\begin{align}
\widetilde{W}_{\sigma_i} = s_i W_{\sigma_i}.
\end{align}
The optimization is then performed on the scaled component. 
Given a learning rate $\eta$ and gradient $\nabla \mathcal{J}(\widetilde{W}_{\sigma_i})$, one update step can be written as
\begin{align}
\widetilde{W}_{\sigma_i}^{t+1}
=
\widetilde{W}_{\sigma_i}^{t}
-
\eta \nabla \mathcal{J}(\widetilde{W}_{\sigma_i}^{t}).
\end{align}
Since the component is optimized in its scaled form, the effective update magnitude is controlled by $s_i$. 
A smaller scaling factor leads to a smaller effective perturbation during optimization, which helps prevent the current task from causing large shifts in the shared singular bases.

This is particularly important because different task-specific components are reconstructed from shared bases $U$ and $V^T$. 
Although each task is assigned to a separated singular component, excessive updates on the current component may still affect the shared basis structure and introduce interference to previously learned subspaces. 
By assigning different scaling factors to different task components, HLM separates their update magnitudes and reduces the chance that one task dominates the shared parameter space.

From an intuitive perspective, suppose the unscaled update magnitude of a task component is bounded by $\Delta$. 
After applying the scaling factor, the effective perturbation can be regarded as being controlled by $s_i \Delta$. 
When $s_i$ is small, this perturbation remains limited, making the shift of shared parameters less likely to exceed an acceptable range. 
Therefore, the scaling mechanism acts as a simple but effective constraint on task-specific updates.

Overall, the scaling mechanism helps HLM maintain the correspondence between incremental tasks and their assigned subspaces, while reducing parameter shifts caused by shared singular bases. 
This provides additional stability during sequential training and contributes to mitigating cross-task subspace interference.



\end{document}